\def\year{2022}\relax
\documentclass[letterpaper]{article} 
\usepackage{aaai22}  
\usepackage{times}  
\usepackage{helvet}  
\usepackage{courier}  
\usepackage[hyphens]{url}  
\usepackage{graphicx} 
\usepackage{natbib}  
\usepackage{caption} 
\DeclareCaptionStyle{ruled}{labelfont=normalfont,labelsep=colon,strut=off} 
\usepackage{algorithm}
\usepackage{algorithmic}

\usepackage{newfloat}
\usepackage{listings}
\floatstyle{ruled}
\newfloat{listing}{tb}{lst}{}
\floatname{listing}{Listing}
\usepackage{booktabs}
\usepackage{amsfonts}
\usepackage{multirow}
\usepackage{pifont}
\usepackage{stfloats}
\usepackage{amsmath}
\usepackage{amsmath, bm}

\title{SoMoFormer: Social-Aware Motion Transformer for Multi-Person Motion Prediction}
\author{
    Xiaogang Peng,\textsuperscript{\rm 1}
    Yaodi Shen,\textsuperscript{\rm 2}
    Haoran Wang,\textsuperscript{\rm 1} 
    Binling Nie,\textsuperscript{\rm 1}
    Yigang Wang,\textsuperscript{\rm 1}
    Zizhao Wu\textsuperscript{\rm 1}
    \thanks{Corresponding author: zizhaowu@hdu.edu.cn}
}
\affiliations{
    \textsuperscript{\rm 1}Hangzhou Dianzi University \\
    \textsuperscript{\rm 2}Beijing University of Posts and Telecommunications \\
    \{pengxiaogang, haoranwang, binlingnie, yigang.wang, zizhaowu\}@hdu.edu.cn, shenyaodi@bupt.edu.cn
}

\usepackage{bibentry}

\begin{document}

\maketitle

\begin{abstract}
Multi-person motion prediction remains a challenging problem, especially in the joint representation learning of individual motion and social interactions. Most prior methods only involve learning local pose dynamics for individual motion (without global body trajectory) and also struggle to capture complex interaction dependencies for social interactions. In this paper, we propose a novel \textbf{So}cial-Aware \textbf{Mo}tion Trans\textbf{former} (\textbf{SoMoFormer}) to effectively model individual motion and social interactions in a joint manner. Specifically, SoMoFormer extracts motion features from sub-sequences in displacement trajectory space to effectively learn both local and global pose dynamics for each individual. In addition, we devise a novel social-aware motion attention mechanism in SoMoFormer to further optimize dynamics representations and capture interaction dependencies simultaneously via motion similarity calculation across time and social dimensions. On both short- and long-term horizons, we empirically evaluate our framework on multi-person motion datasets and demonstrate that our method greatly outperforms state-of-the-art methods of single- and multi-person motion prediction. Code will be made publicly available upon acceptance. 
\end{abstract}

\section{1\quad Introduction}
Recent years have seen a proliferation of work on the topic of human motion prediction \cite{c:7,c:15,c:19,c:20,c:22,c:24}, which aims to forecast future poses based on past observations. Understanding and forecasting human motion also plays a critical role in the field of artificial intelligence and computer vision, especially for robot planning, autonomous driving, and video surveillance \cite{c:21,c:25,c:26,c:16}.

Although encouraging progress has been achieved, the current methods are mostly based on local pose dynamics forecasting without global joints' position changes (global body trajectory) and often tackle the problem for single humans in isolation without human-human interaction. However, in real-world scenarios, each person may interact with one or more people, ranging from low to high levels of interactivity with instantaneous and deferred mutual-influences \cite{b:1, c:32}. 
\begin{figure}[t]
\centering
\includegraphics[width=0.45\textwidth]{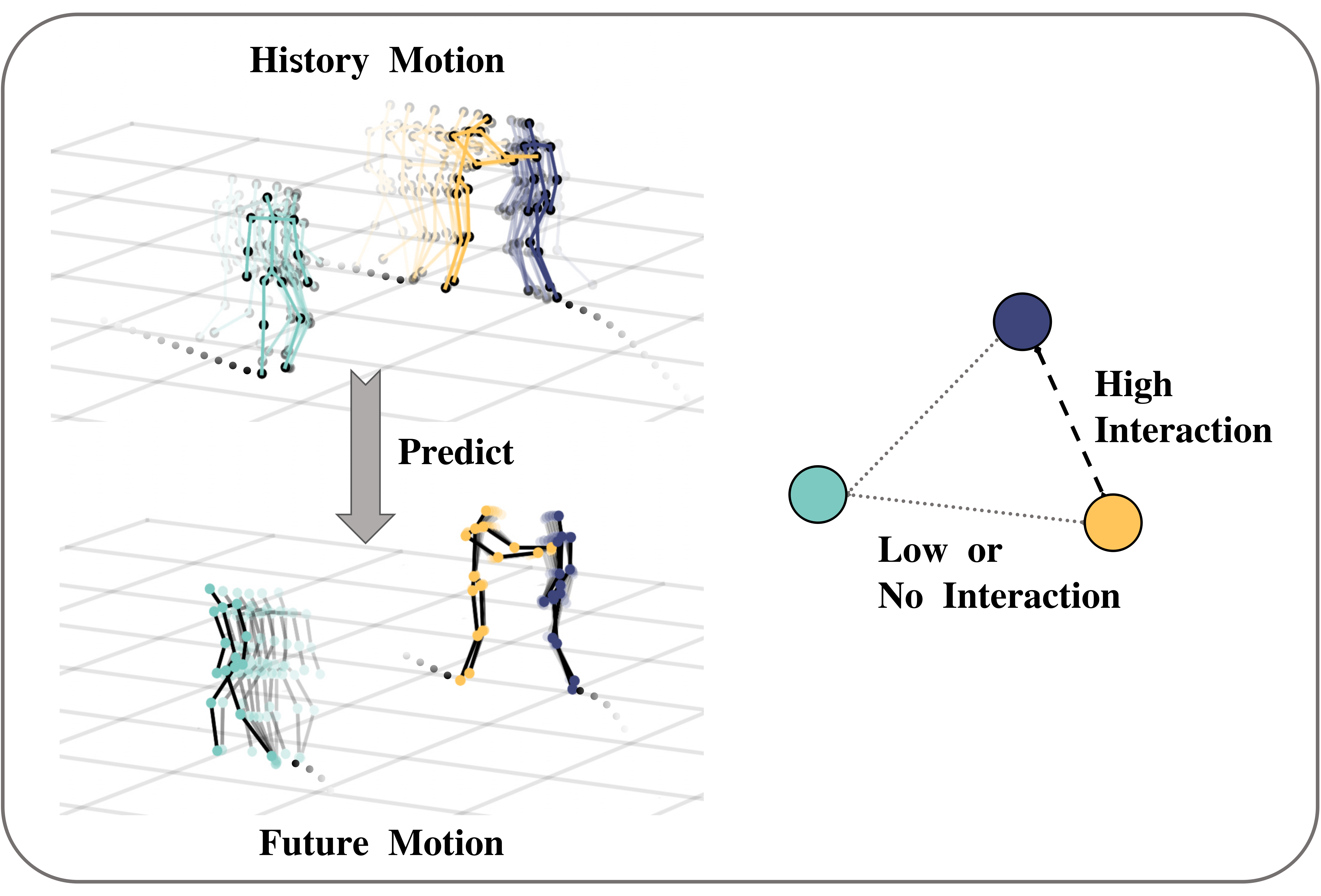} 
\caption{A complex scene of multi-person motion should contain individuals with different levels of interaction (low and high interaction). The proposed framework allows for modeling such a complex context effectively.}
\label{fig1}
\end{figure}
For example, Figure \ref{fig1} gives us an illustration, where two persons are pushing and shoving with high interaction, while another person is just walking on the side with no or low interaction. Thus, accurately forecasting pose dynamics and trajectory for individual motion and comprehensively considering complex interactive factors for social interactions are imperative for understanding human behavior in multi-person motion prediction. However, existing solutions do not jointly consider these challenging factors in an efficient way. For instance, Guo \textit{et al.} \cite{c:14} first propose a collaborative prediction task and perform future motion prediction for only two interacted dancers, which inevitably ignores low interaction influence on one’s future behavior. Besides, in the scene with more individuals, Wang \textit{et al.} \cite{c:3} use local-global Transformers to learn individual motion and social interactions separately followed by feature concatenation. In this work, we argue this strategy is sub-optimal since individuals' behavior can be influenced by themselves and others at different times. Additionally, they take a vanilla Transformer as the local Transformer for temporal modeling, which neglects the spatial dependencies of the human body joints.

To address these challenges, we propose a novel Transformer-based framework, termed SoMoFormer, which contains a displacement sub-sequence encoder (DSE), a social interaction encoder (SIE), and a Transformer predictor. Firstly, DSE transforms observed sequences to displacement sub-sequences and extracts motion features from the sub-sequences via multiple graph convolutional network (GCN) units to faithfully learn local and global pose dynamics. Then we use MLP to downsample the motion features for each individual, which will further facilitate the learning of pose dynamics and social interactions compared to frame-wise pose features.

Secondly, the Transformer-based SIE that comprises a time encoder, a spatial encoder, and a social-aware motion attention mechanism (SAMA), is presented to jointly model individual motion and social interactions. Since Transformer requires a single sequence as input, the SIE uses a multi-person displacement sequence that is flattened by the sub-sequences of each person across time and individuals. The time encoder applies temporal positional encoding to such a sequence representation to preserve temporal information for each person. The spatial encoder produces spatial relations between individuals and supplements them in the SAMA mechanism as attention biases. The SAMA mechanism allows each person to attend to different inter- and intra-individual motion relations across time and social dimensions. This mutual consideration of relationships in time and space based on motion features rather than pose features can further optimize dynamic representations and effectively capture various interaction dependencies for complex social interactions.

Finally, a Transformer decoder (dubbed the Transformer predictor) is introduced to further consider the relations between the current and historical context across individuals and predict smooth and accurate multi-person motion trajectories. For multi-person motion prediction (with 3 persons), we evaluate our method on multiple datasets, including CMU-Mocap \cite{c:1} with UMPM \cite{c:5} augmented and MuPoTS-3D \cite{c:4}. Besides, we extend our experiment by mixing the above datasets with the 3DPW \cite{c:2} dataset to perform prediction in a more complex scene (with 6 $\sim$ 10 persons). Our method outperforms the state-of-the-art approaches for both short- and long-term predictions by a large margin, with 12 $\sim$ 36$\%$ accuracy improvement for the short term ($\le$ 0.4s) and 8 $\sim$ 16$\%$ accuracy improvement for the long term(0.4s $\sim$ 1.0s).

To summarize, our key contributions are as follows: 
\begin{itemize}
    \item We propose a novel Transformer-based framework that effectively models individual motion and social interactions in a joint manner for plausible multi-person motion prediction. 
    \item We introduce a displacement sub-sequence encoder to faithfully learn local and global pose dynamics from sub-sequences in the displacement trajectory space.
    \item We present a novel social-aware motion attention mechanism that simultaneously considers inter- and intra-individual motion relations to further optimize dynamics representations and capture different interaction dependencies.
    \item On multiple multi-person motion datasets, the proposed SoMoFormer significantly outperforms the state-of-the-art methods of single- and multi-person motion prediction.
\end{itemize}

\section{2\quad Related Work}
\subsubsection{Human Motion Prediction.}Predicting human motion offers enormous promise for surveillance, autonomous driving, and human-robot interaction. Although recurrent neural networks (RNNs) have shown advantages in processing this typical sequence-to-sequence problem \cite{c:15,c:36,c:37}, discontinuity and error accumulation often happen due to the frame-by-frame prediction manner. To address these issues, some feed-forward networks such as graph convolution networks (GCNs) and temporal convolution networks (TCNs) are used to explore spatial and temporal dependencies \cite{c:19,c:20,c:29,c:31}. Besides, Mao \textit{et al.} \cite{c:12} introduce an attention-based feed-forward network to capture the similarity between the current motion context and the historical motion sub-sequences and process the result with a GCN for long-term prediction. All the aforementioned methods fix the body center position for local pose dynamics forecasting, thus ignoring global body trajectory. In addition, interactions at the skeleton level between individuals are not modeled or captured. However, we argue that learning both local and global pose dynamics simultaneously and modeling fine-grained human-human interactions are essential for understanding complex human behavior. Different from prior work, our SoMoFormer can effectively capture both local and global pose dynamics in displacement trajectory space and model human-human interaction at a skeleton level.

\begin{figure*}[t]
\centering
\includegraphics[width=1.0\textwidth]{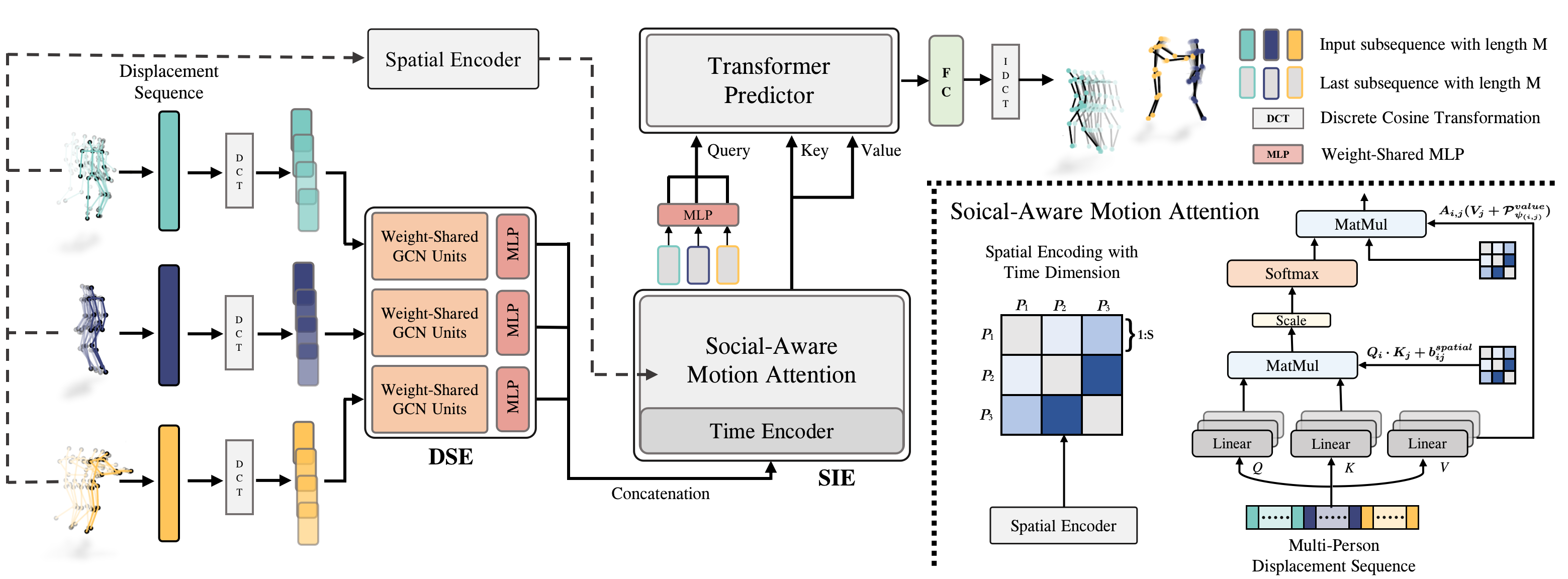} 
\caption{Left: Overview of our SoMoFormer framework. Right: Illustration of social-aware motion attention and spatial encoding.}
\label{fig2}
\end{figure*}

\subsubsection{Multi-Person Social Interaction}Multi-person trajectory prediction is a representative issue for social interaction. Existing methods for the task can be categorized based on how they model time and social dimensions. RNNs \cite{c:40} and Transformers \cite{c:41} are the preferred models \cite{c:42,c:43,c:44} to process the trajectory sequence for temporal modeling, and graph neural networks (GNNs) \cite{c:45} are often adopted as social models for interaction modeling \cite{c:46,c:47}. While performing well, these studies only focus on individuals' global movement without modeling detailed human joint dynamics. In order to address fine-grained human-human interaction, some approaches are proposed to predict multi-person motion trajectories. For example, Adeli \textit{et al.} \cite{c:13} propose to combine scene context for multi-person motion prediction on a short-term horizon, and Guo \textit{et al.} \cite{c:14} present a collaborative prediction task for only two interacted persons. Furthermore, Wang \textit{et al.} \cite{c:3} introduce a Transformer-based framework to forecast motion trajectory for more people. Despite the novelty of these methods, individual motion and social interaction are often modeled separately without effective feature fusion, rendering them incapable of handling different levels of interaction in crowd scenarios. In this work, we investigate our SoMoFormer to model individual motion and social interactions simultaneously and predict future motion for 3 $\sim$ 10 persons with complex social interactions.


\section{3\quad Method}
As shown in Figure \ref{fig2}, the overall framework comprises a displacement sub-sequence encoder (DSE), a social interaction encoder (SIE), and a Transformer predictor. Besides, we adopt a Discrete Cosine Transformation (DCT) that discards the high-frequency information for a more compact representation in trajectory space \cite{c:9,c:10}. In the following, we introduce the problem definition and our key modules in detail.

\subsection{3.1\quad Problem Definition}
 Supposing the historical poses from person $p$ are $X_{1:N}^p=\{x_1^p,x_2^p,...,$ $x_N^p\}$ with $N$ frames, where $p = 1,2, ... P$. For simplicity, we omit subscript $p$ when $p$ only represents an arbitrary person. e.g., taking $x^p_{1:t}$ as $x_{1:t}$. Instead of absolute joint positions in the world coordinate, we use ${y}_i = x_{i+1} - x_{i}$ to obtain instantaneous pose displacement at time $i$, which provides more valuable dynamics information \cite{c:7}. The whole displacement sequence is defined as $Y_{1:N-1} = \{{y_1,y_2,...,y_{N-1}}\}$. Given the displacement sequence $Y_{1:N-1}$ from each person, our goal is to predict the future displacement trajectory $Y_{N:N+T-1}$ and transform it back to the pose space $X_{{N+1}:N+T}$.

\subsection{3.2\quad Displacement Sub-sequence Encoder (DSE)}
Due to that humans tend to repeat their motion across a period of time \cite{c:12}, we divide the displacement sequence of each person $Y_{1:N-1} = \{{y_1,y_2,...,y_{N-1}}\}$ into $N-M$ sub-sequences $\{Y_{i:i+M-1}\}_{i=1}^{N-M}$. To faithfully learn local and global pose dynamics, we exploit multiple GCN units (Multi-GCN Units) to directly extract features of human body dynamics from the sub-sequences in displacement trajectory space. Specifically, each of the unit is composed of one initial GCN, one end GCN, and 2 residual GCNs. Note that the above unit with GCNs is similar to the method \cite{c:9} except for the number of residual GCNs. In addition, due to the limitation of GCNs in the time dimension, it is hard to handle sequences of an arbitrary length during training. To solve this, we use the length of the sub-sequence $M$ as the kernel size and separate the first $S$ sub-sequences into the Mutli-GCN Units with $S$ units to explore individual's spatial dependencies among distinct joints, where $S=N-M-1$. The output $\tilde{Y}\in \mathbb{R}^{BPS \times M \times F}$ of the Multi-GCN Units is then fed into MLP to downsample each sub-sequence from length $M$ to $1$, where $BPS$ is the merged dimension size and $F$ is the feature dimension. Formally,
\begin{equation}\label{eq1}
\tilde{Y} \xrightarrow[convolution]{1\times 1} \tilde{Y}'\in{\mathbb{R}^{BPS \ \times F}},
\end{equation}
\begin{equation}\label{eq2}
LeakyRelu(\tilde{Y}') \xrightarrow[]{reshape} \tilde{Y}''\in{\mathbb{R}^{B\times P\times S \times F}}.
\end{equation}
The last observed sub-sequence from each person is taken as a query motion for the Transformer predictor, which we shall discuss later in subsection 3.4.

\subsection{3.3\quad Social Interaction Encoder (SIE)}
Inspired by Transformers in the graph domain \cite{c:8,c:18,c:33,c:34,c:35}, we introduce a social interaction encoder (SIE) based on Transformer to model individual motion and social interactions simultaneously. In this subsection, we illustrate the details of the SIE from the following four components: multi-person displacement sequence, time encoder, spatial encoder, and social-aware motion attention. 

\subsubsection{Multi-Person Displacement as a Sequence.} After the operation by DSE, individuals' sequence is represented as $\tilde{Y}''\in{\mathbb{R}^{B\times P\times S \times F}}$. We reshape all the sequences and denote past multi-person displacement sequence as $Y_{past}=(y_{1}^{1},$  $...,y_{S}^{1},...,y_{1}^{P},...,y_{S}^{P}) \in{\mathbb{R}^{B\times PS \times F}}$ of length $L = P \times S$, which are used for the Transformers below.

\subsubsection{Time Encoder.} We employ a time encoder proposed by \cite{c:30} to preserve individuals' temporal information in the multi-person displacement sequence. The time encoder takes $Y_{past}$ as input to calculate timestamp features based on the element's timestamp and uses the same sinusoidal design from the original Transformer as the positional encoding on the timestamp.

\begin{figure}[t]
\centering
\includegraphics[width=0.45\textwidth]{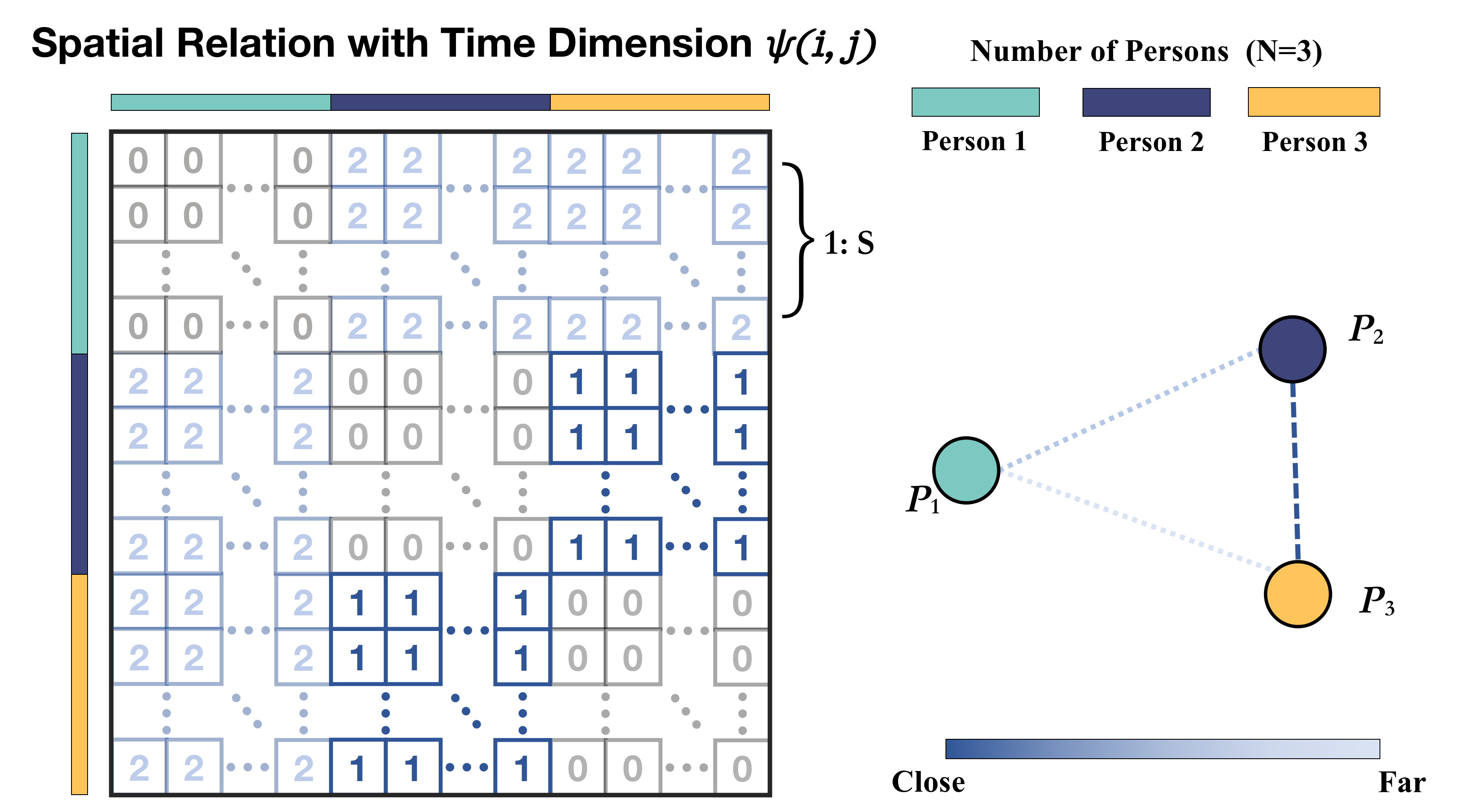} 
\caption{The illustration of spatial relation with the time dimension between different individuals. S is the individuals' sequence length.}
\label{fig3}
\end{figure}

\subsubsection{Spatial Encoder.}In practice, the closer people are to each other in 3D space, the higher their interactivity will be to a certain extent. Here, we propose a spatial encoder to capture such intangible connections among people. Specifically, we first extract the root (central hip) position sequence from the original pose space, and then segment and downsample it as DSE does. To share temporal information, we also flatten the sequences across time and individuals to obtain a root position sequence. Given the sequence, we compute the Euclidean distance across time and individuals to obtain a 3D relative position matrix: $e_{i,j} = \sqrt{(\tilde{x}_i - \tilde{x}_j)^2+(\tilde{y}_i - \tilde{y}_j)^2+(\tilde{z}_i - \tilde{z}_j)^2},  $ where $i,j = 1,..,L$. 

Next, we map this relative position matrix to an integer set for a spatial relation. In our opinion, one individual can be affected to some extent by others from little far away. However, the widely used clip function $h(e)=max(-\beta,min(\beta, e))$ eliminates the context of long-distance relative position. Inspired by \cite{c:11},  we use the piecewise index function to get spatial relation $\psi_{(i,j)}$ with the time dimension as shown in Figure \ref{fig3}, containing information both in the short- and long-range. The piecewise index function is presented as
\begin{equation}\label{eq3}
\small{
g(e)=\left\{
\begin{aligned}
[e],\quad\quad\quad\quad\quad\quad\quad\quad\quad\quad\quad\quad\quad\quad\quad\quad\quad |e|\le\alpha\\
sign(e)\times min(\beta, [\alpha+\frac{\ln{(|e|/\alpha)}}{\ln{(\gamma/\alpha)}} (\beta-\alpha)]), |e|>\alpha\\
\end{aligned}
\right.
}
\end{equation}
where $[\cdot]$ is a round operation, $sign()$ determines the sign of a number, i.e., returning 1 for positive input, -1 for negative, and 0 for otherwise. $\alpha$ controls the piecewise point, $\beta$  limits the output in the range of $[- \beta, \beta]$, and $\gamma$ tunes the curvature of the logarithmic part. 

Finally, we embed the spatial relation as spatial encoding for query, key, and value respectively: $ \mathcal{P}^{query}, \mathcal{P}^{key},$  $\mathcal{P}^{value}\in \mathbb{R}^{L\times d_z}$, which are shared throughout all attention layers.

\subsubsection{Social-Aware Motion Attention.}We aim to construct a social-aware motion attention (SAMA) mechanism to effectively model individual motion and complex social interactions in a joint manner, as shown in Figure \ref{fig2}-right. With ample motion features acquired by downsampling from the sub-sequences, SAMA, based on motion-wise attention computation, can further optimize pose dynamics representations and capture different interaction dependencies among individuals by measuring motion similarity rather than pose similarity. Let $H=[h_1^{},...,h_n]\in \mathbb{R}^{n\times d}$ denote the input representation for attention module, where $d$ is the hidden dimension. SAMA takes as input keys $K$, queries $Q$ and values $V$, each of which is projected by the corresponding parameter matrix $W_Q\in \mathbb{R}^{d\times d_z}$, $W_K\in \mathbb{R}^{d\times d_z}$ and $W_V\in R^{d\times d_z}$. The output of social-aware motion attention is computed as
\begin{equation}\label{eq4}
Q=HW_Q,\quad K=HW_K,\quad V=HW_V,
\end{equation}
\begin{equation}\label{eq5}
{\rm SAMA}(Q,K,V) = softmax(A)\tilde{V}.
\end{equation}

Inspired by \cite{c:6}, we integrate the spatial encoding on the attention map, which considers the interaction between individual features and spatial relations across time in this social interaction graph. Denoting $A_{ij}$ as the $(i,j)$-element of the Query-Key product matrix $ A $, we have 
\begin{equation}\label{eq6}
A_{ij} = \frac{Q_{i} \cdot K_j+ b_{i,j}^{spatial}}{\sqrt{d_z}}, 
\end{equation}

\begin{equation}\label{eq7}
b_{ij}^{spatial} = Q_{i} \cdot  \mathcal{P}^{query}_{\psi_{(i,j)}} + K_{i} \cdot  \mathcal{P}^{key}_{\psi_{(i,j)}}.
\end{equation}

The advantage of such an operation is that compared to conventional GNNs with a restricted receptive field, the Transformer layer can provide more global information \cite{c:8}. Besides, each individual in a single Transformer layer can adaptively attend to itself and other individuals across time and social dimensions based on spatial and motion features. To maintain the preciseness of spatial information, we encode individuals into the hidden features of values:

\begin{equation}\label{eq8}
    \tilde{V_j} = V_j+\mathcal{P}^{value}_{\psi_{(i,j)}}. 
\end{equation}


\begin{table*}[t]
\centering
\resizebox{\linewidth}{!}{
\begin{tabular}{@{}clcccccccccccccccccccc@{}}
\toprule
\multicolumn{1}{l}{} & \multicolumn{1}{c}{} & \multicolumn{5}{|c}{\begin{tabular}[c]{@{}c@{}}CMU-Mocap (UMPM)\\ (3 persons)\end{tabular}} & \multicolumn{5}{|c}{\begin{tabular}[c]{@{}c@{}}MuPoTS-3D\\ (3 persons)\end{tabular}} & \multicolumn{5}{|c}{\begin{tabular}[c]{@{}c@{}}Mix1\\ (6 persons)\end{tabular}} & \multicolumn{5}{|c}{\begin{tabular}[c]{@{}c@{}}Mix2\\ (10 persons)\end{tabular}} \\ \midrule
                     & Time (sec) & \multicolumn{1}{|c}{0.2} & 0.4 & 0.6 & 0.8 & 1.0 & \multicolumn{1}{|c}{0.2}& 0.4 & 0.6 & 0.8 & 1.0 & \multicolumn{1}{|c}{0.2} & 0.4 & 0.6 & 0.8 & 1.0 & \multicolumn{1}{|c}{0.2} & 0.4 & 0.6 & 0.8 & 1.0 \\
                     \midrule
\multirow{4}{*}{\rotatebox{90}{\textbf{MPJPE}}} 
& Hisrep (ECCV 2020) & \multicolumn{1}{|c}{49} & 92 & 130 & 164 & 207 &  \multicolumn{1}{|c}{62} & 111 & 157 & 192 & 242 &  \multicolumn{1}{|c}{51} & 98 & 141 & 184 & 233 &  \multicolumn{1}{|c}{52} & 99 & 140 & 179 & 224 \\
& MSR (ICCV 2021)  & \multicolumn{1}{|c}{45} & 100 & 146 & 189 & 231 &  \multicolumn{1}{|c}{60} & {108} & {153} & \textbf{195} & \textbf{239} &  \multicolumn{1}{|c}{44} & 85 & 127 & 169 & 211 &  \multicolumn{1}{|c}{60} & 106 & 153 & 197 & 243 \\
& MRT*   (NeurIPS 2021) & \multicolumn{1}{|c}{36} & {76} & {115} & {152} & {192} &  \multicolumn{1}{|c}{58} & 114 & 169 & 216 & 267 &  \multicolumn{1}{|c}{37} & {78} & {122} & \textbf{165} & {212} &  \multicolumn{1}{|c}{38} & {82} & {126} & {168} & {214} \\
& Ours*  & \multicolumn{1}{|c}{\textbf{34}} & \textbf{74} & \textbf{112} & \textbf{149} & \textbf{186} &  \multicolumn{1}{|c}{\textbf{50}} & \textbf{101} & \textbf{151} & {196} & {241} &  \multicolumn{1}{|c}{\textbf{36}} & \textbf{77} & \textbf{122} & \textbf{165} & \textbf{211} &  \multicolumn{1}{|c}{\textbf{36}} & \textbf{79} & \textbf{122} & \textbf{163} & \textbf{202} \\ \midrule

\multirow{4}{*}{\rotatebox{90}{\textbf{AMPJPE}}} 
& Hisrep (ECCV 2020) & \multicolumn{1}{|c}{41} & 71 & 97 & 113 & 130 &  \multicolumn{1}{|c}{56} & 90 & 118 & 138 & 155 &  \multicolumn{1}{|c}{38} & 66 & 92 & 107 & 122 &  \multicolumn{1}{|c}{41} & 73 & 100 & 117 & 133 \\
& MSR (ICCV 2021) & \multicolumn{1}{|c}{40} & 71 & 94 & 112 & 126 &  \multicolumn{1}{|c}{56} & {90} & {116} & 136 & 152 &  \multicolumn{1}{|c}{37} & 65 & 87 & 103 & 116 &  \multicolumn{1}{|c}{48} & 81 & 110 & 132 & 148 \\
& MRT*  (NeurIPS 2021)  & \multicolumn{1}{|c}{36} & {74} & {108} & {132} & {159} &  \multicolumn{1}{|c}{56} & 101 & 138 & 162 & 188 &  \multicolumn{1}{|c}{36} & {74} & {109} & 134 & {166} &  \multicolumn{1}{|c}{38} & {78} & {115} & {143} & {178} \\
& Ours*  & \multicolumn{1}{|c}{\textbf{29}} & \textbf{61} & \textbf{87} & \textbf{108} & \textbf{123} &  \multicolumn{1}{|c}{\textbf{47}} & \textbf{84} & \textbf{113} & {\textbf{135}} & {\textbf{151}} &  \multicolumn{1}{|c}{\textbf{31}} & \textbf{62} & \textbf{85} & \textbf{101} & \textbf{115} &  \multicolumn{1}{|c}{\textbf{33}} & \textbf{69} & \textbf{94} & \textbf{112} & \textbf{128} \\

\bottomrule
\end{tabular}
}
\caption{Results of MPJPE and AMPJPE (in mm) on different datasets. We compare our method with the previous SOTA methods for short-term and long-term predictions. Best results are shown in boldface. (* means multi-person motion prediction method.)}
\label{table1}
\end{table*}

\subsection{3.4\quad Transformer Predictor}
As illustrated in Figure \ref{fig2}-left, the last observed sub-sequence from each person (Note that it is not through the DSE) is input to MLP for downsampling. Then we reshape all of them into one multi-person displacement sequence and take it as the queries.  Keys and values are provided by SIE’s output. We utilize widely-used multi-head attention in the Transformer predictor to consider the relations between the current (queries) and historical context (keys) across individuals. At the end of the predictor, we adopt two fully connected layers followed by an Inverse Discrete Cosine Transformation (IDCT) \cite{c:10} to generate the future motion trajectory $X_{{N+1}:N+T}$ for each individual.

\subsection{3.5\quad Loss Function}
We use a reconstruction loss based on the Mean Per Joint Position Error (MPJPE). In particular, for one training sample, the loss is represented as
\begin{equation}\label{eq9}
L_{rec} = \frac{1}{J*T}\sum_{i=N+1}^{N+T}{\sum_{j=1}^{J}{||\hat{y}_{i,j} - y_{i,j}||^2}},
\end{equation}
where $\hat{d}_{j,t}$ and $d_{j,t}$ are ground-truth and estimated pose displacement at time $i$. $J$ is the number of body joints.

\subsection{3.6\quad Implementation Details}
We implement our framework in PyTorch, and the experiments are performed on Nvidia GeForce RTX 3090 GPU. We train our model for 50 epochs using the ADAM optimizer with a batch size of 32, a learning rate of 0.0003, and a dropout of 0.2. For the DSE, the kernel size of GCNs is 10, equal to the sub-sequence length, and $F=128, D=3J$, where $J=15$ is the number of skeleton joints. The parameters of the piecewise index function in spatial encoder are: $\alpha=1,\beta=2,\gamma=4$. The dimensions $d_z$ of keys, queries, and values in SIE and Transformer predictor are all set to 64, and the hidden dimension $d$ of feedforward layers is 1024. The SIE and Transformer predictor both have 3 stacked attention layers with 8 heads. All MLPs in the DSE and spatial encoder have the hidden dimensions ($D$, $F$) and (3, 3) respectively. During training, we adopt the strategy in the method proposed by \cite{c:3} to predict future motion recursively for a longer horizon.

\section{4\quad Experiments}
\subsection{4.1\quad Datasets}
To verify the effectiveness of SoMoFormer, we run experiments on CMU-Mocap (UMPM) dataset, which is augmented and enlarged by UMPM \cite{c:5} on the CMU-Mocap \cite{c:1} dataset. Mix1 and Mix2 are blended by CMU-Mocap, UMPM, 3DPW \cite{c:2}, and MuPoTs-3D \cite{c:4} datasets. We evaluate all the methods for generalization ability by testing on the MuPoTS-3D (3 persons), Mix1 (6 persons), and Mix2 (10 persons) datasets with the model only trained on the CMU-Mocap (UMPM) dataset.

\subsection{4.2\quad Metrics of Evaluation}

\subsubsection{MPJPE.} The most widely used metric for pose estimation and motion prediction tasks is the mean per joint position error (MPJPE). Here we use the MPJPE to measure the error of individuals' poses, including body trajectory:
\begin{equation}\label{eq10}
{\rm MPJPE}(P,G)=\frac{1}{N\times J}\sum_{i=1}^{N}\sum_{j=1}^{J}{||P_j^i - G_j^i||^2},
\end{equation}
where $N$ and $J$ are the numbers of people and joints. $P_j^i$ and $G_j^i$ are the estimated and ground-truth positions of the joint $j$ for person $i$.

\begin{figure}[h]
\centering
\includegraphics[width=0.45\textwidth]{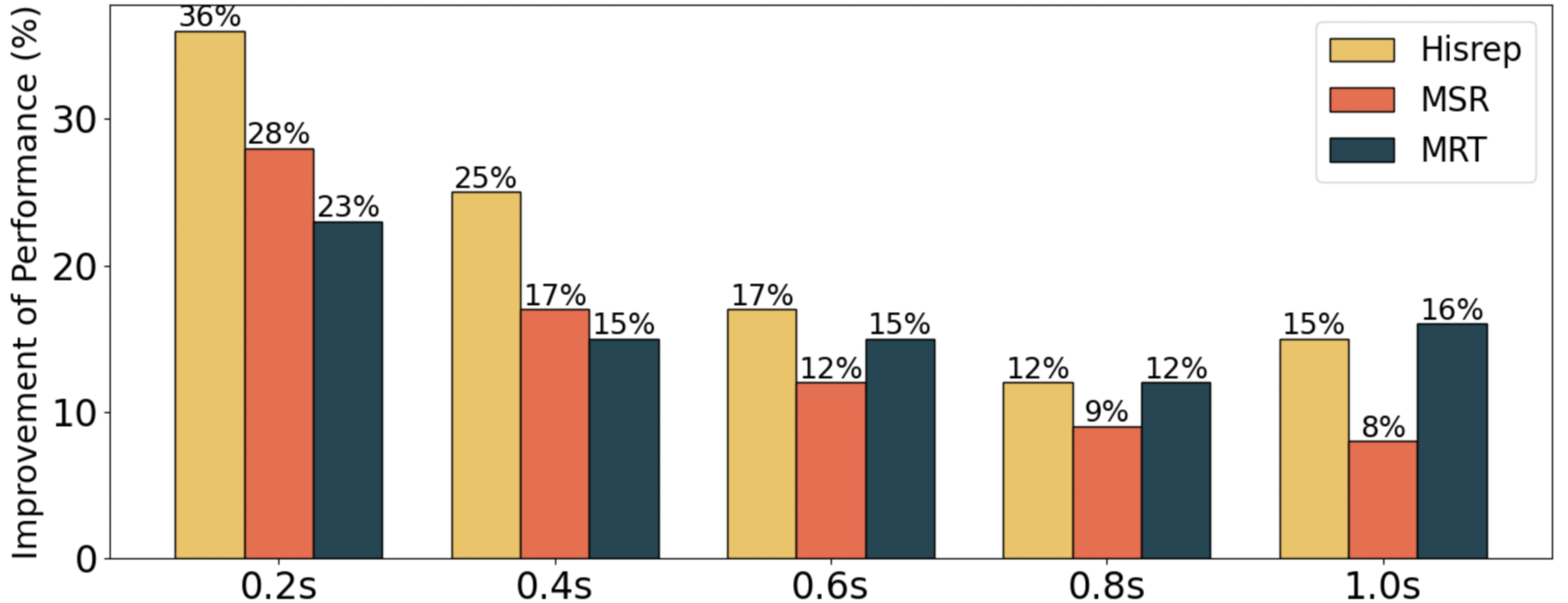} 
\caption{Average improvement of performance in the 4 metrics compared with the baselines.}
\label{fig4}
\end{figure}

\subsubsection{AMPJPE.} We remove global movement and use aligned MPJPE to measure pure pose position error:
\begin{equation}\label{eq11}
{\rm AMPJPE}(P,G)={\rm MPJPE}(P-P_{r},G-G_{r}),
\end{equation}
where $P_{r}$ and $G_r$ are the estimated and ground-truth root positions.

\subsubsection{ADE and FDE.} We also take the root position to evaluate the global movement of each person using typical trajectory prediction metrics. The formulas are described as follows:
\begin{equation}\label{eq12}
{\rm ADE}(P,G) = \frac{1}{T}\sum_{t=1}^{T}{||P_{r,t} - G_{r,t}||^2},
\end{equation}
\begin{equation}\label{eq13}
{\rm FDE}(P,G) = {||P_{r,T} - G_{r,T}||^2},
\end{equation}
where $P_{r,T}$ and $G_{r,T}$ are the estimated and ground-truth root position of final pose at timestamp $T$.

\subsection{4.3\quad Baselines}
We choose 3 code-released state-of-the-art (SOTA) approaches as baselines, including two single-person based methods: Hisrep \cite{c:12} and MSR \cite{c:29}, and a recently released multi-person based method called MRT \cite{c:3}, which both allow absolute coordinates as input. For a fair comparison, all these models are trained with 50 frames (2.0 s) of input and 25 frames (1.0 s) of forecasting and evaluated on the 4 datasets.

\subsection{4.4\quad Results }
To validate the prediction performance of SoMoFormer, we follow the setting of the most single-person methods \cite{c:12,c:29} to show the quantitative and qualitative results of short-term (i.e., 10 frames) and long-term (i.e., 25 frames) predictions and compare our method with the baselines.

\begin{figure*}[t]
\centering
\includegraphics[width=0.9\textwidth]{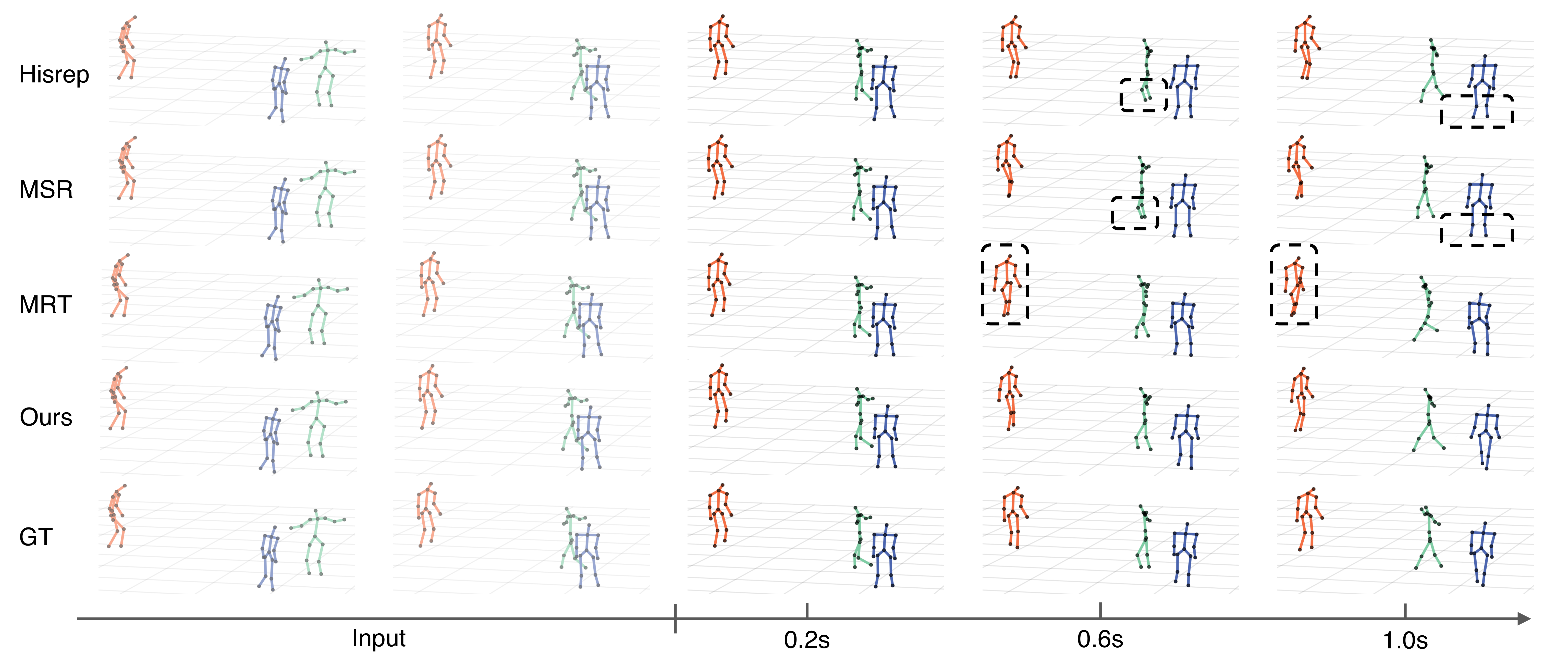} 
\caption{Qualitative comparison with the baselines and the ground truth on a sample of the CMU-Mocap (UMPM) dataset. The left two columns are inputs, and the right three columns are predictions.}
\label{fig5}
\end{figure*}

\subsubsection{Quantitative Results.}Table \ref{table1} reports the results of MPJPE and AMPJPE on the 4 different datasets. We can observe that our SoMoFormer significantly outperforms the baselines in prediction accuracy. It is worth mentioning that our method surpasses the baselines by a small margin, despite much noise in the MuPoTs-3D dataset. Moreover, it is clear that MRT performs poorly in the AMPJPE metric as a result of lacking the spatial modeling of the human skeleton. The average performance improvement in the 4 metrics compared with the baselines is shown in Figure \ref{fig4}.

\begin{table}[h]
\resizebox{\linewidth}{!}{
\begin{tabular}{@{}cccccccccc@{}}
\toprule
$\mathcal{D}$ & DSE & SIE & $\mathcal{P}_{\psi_{(i,j)}}$  & I/DCT & \multicolumn{1}{|c}{0.2} & 0.4 & 0.6 & 0.8 & 1.0 \\ \midrule
& \ding{51}   & \ding{51} &  \ding{51}  & \ding{51} &\multicolumn{1}{|c}{175}  & 196  & 227 & 617 & 621 \\
\ding{51} &   & \ding{51} & \ding{51}  &\ding{51}  &  \multicolumn{1}{|c}{33}  & 77  & 119 & 160 & 200 \\
\ding{51}  &  \ding{51}   &    &   &\ding{51}  &  \multicolumn{1}{|c}{37}   & 75   & 114   & 152   & 192   \\
\ding{51}  & \ding{51}  & \ding{51}   &  & \ding{51}  &  \multicolumn{1}{|c}{36}  & 75  & 113 & 150 & 189 \\
\ding{51}  & \ding{51}  & \ding{51} & \ding{51} &    & \multicolumn{1}{|c}{41}  & 83  & 122 & 159 & 195 \\
\ding{51}  & \ding{51}  & \ding{51} & \ding{51}  &\ding{51}  & \multicolumn{1}{|c}{\textbf{34}}  & \textbf{74}  & \textbf{112} & \textbf{149} & \textbf{186} \\ \bottomrule
\end{tabular}
}
\caption{Ablation studies on different components of SoMoFormer. Our method and its variants are evaluated on the CMU-Mocap (UMPM) in MPJPE metric.}
\label{table3}
\end{table}

\subsubsection{Qualitative Results.} Figure \ref{fig5} shows some examples of our visualization results compared to the baselines and the ground truth. It can be seen that our predictions are much closer to the ground truth than the other methods. The results of Hisrep \cite{c:12} and MSR \cite{c:29} show that they tend to converge to a static pose in long-term predictions. The result of MRT \cite{c:3} shows some pose distortion due to the deficiency in individual motion modeling.

\subsection{4.5\quad Ablation Studies}
We further perform extensive ablation studies on CMU-Mocap (UMPM) to investigate the contribution of key technical components in our method, with results in Table \ref{table3}.

\subsubsection{Effects of DSE.} The goal of the DSE is to extract motion features from displacement sub-sequences. As shown in the first row of Table \ref{table3}, where we directly process input sequences from pose space rather than displacement trajectory space $\mathcal{D}$, the prediction accuracy is greatly affected. As illustrated in the second row, where the DSE is removed, our framework only extracts pose features in a frame-wise manner, resulting in a sub-optimal performance for long-term predictions. This validates our idea that motion features extracted from sub-sequences in displacement trajectory space will boost learning of pose dynamics and social interactions. 

\begin{table}[h]
\resizebox{\linewidth}{!}{
\begin{tabular}{@{}lccccc@{}}
\toprule
Different Sub-sequence Processing  & \multicolumn{1}{|c}{0.2} & 0.4 & 0.6 & 0.8 & 1.0 \\ \midrule
sub-sequence length = 1  & \multicolumn{1}{|c}{35} & 78  & 120 & 162 & 202 \\
sub-sequence length = 5  & \multicolumn{1}{|c}{\textbf{34}}  & 75  & 115 & 153 & 200 \\
sub-sequence length = 10 (ours)  & \multicolumn{1}{|c}{\textbf{34}}  & \textbf{74}  & \textbf{112} & \textbf{149} & \textbf{186} \\
sub-sequence length = 15  &  \multicolumn{1}{|c}{36}  & 76  & 114 & 152 & 188 \\ \midrule
dividing stride = 1 (ours) & \multicolumn{1}{|c}{\textbf{34}}  & \textbf{74}  & \textbf{112} & \textbf{149} & \textbf{186} \\
dividing stride = 3  &  \multicolumn{1}{|c}{35} & 75  & 113 & 152 & 191 \\
dividing stride = 5  &  \multicolumn{1}{|c}{35}  & 75  & 115 & 153 & 190 \\ 
dividing stride = 7  &  \multicolumn{1}{|c}{36}  & 76  & 114 & 150 & 188 \\ \bottomrule
\end{tabular}
}
\caption{Ablation studies on different lengths and dividing strides of sub-sequence with MPJPE metric.}
\label{table4}
\end{table}

\subsubsection{Effects of SIE.} The SIE is aimed at modeling individual motion and complex social interactions simultaneously. When it is eliminated, which means that the DSE's output is delivered directly to the Transformer predictor for decoding, the performance drops substantially. As shown in the fourth row, where we keep main part of the SIE and only remove all the spatial encodings $\mathcal{P}_{\psi_{(i,j)}}$ in the attention module, the performance still could not be optimal. 

Besides, we remove the DCT and IDCT (I/DCT) at the framework’s head and tail, which drastically reduces performance. In conclusion, we can see that all variants result in inferior performance compared to our full method.

\subsubsection{Effects of Different Sub-sequence Processing Manners.} Different processing manners affect the richness of the motion features acquired during learning. Thus, we also conduct ablations on different sub-sequence processing manners with the results in Table \ref{table4}. As shown, our default processing produces better average results.



\section{Conclusion}
In this paper, we introduce a novel Transformer architecture to predict multi-person motion with complex social interactions. We first present a displacement sub-sequence encoder for learning both local and global pose dynamics in displacement trajectory space. Then, a Transformer equipped with a social-aware motion attention mechanism is proposed, which effectively models individual motion and social interactions in a joint manner. Finally, we propose a Transformer predictor to generate plausible future motion trajectory. Experiments demonstrated that our method significantly improved state-of-the-art performance on the different multi-person motion datasets.


\section{References}
\nobibliography*
\bibentry{c:1}.\\[.2em]
\bibentry{c:2}.\\[.2em]
\bibentry{c:3}.\\[.2em]
\bibentry{c:4}.\\[.2em]
\bibentry{c:5}.\\[.2em]
\bibentry{c:6}.\\[.2em]
\bibentry{c:7}.\\[.2em]
\bibentry{c:8}.\\[.2em]
\bibentry{c:9}.\\[.2em]
\bibentry{c:10}.\\[.2em]
\bibentry{c:11}.\\[.2em]
\bibentry{c:12}.\\[.2em]
\bibentry{c:13}.\\[.2em]
\bibentry{c:14}.\\[.2em]
\bibentry{c:15}.\\[.2em]
\bibentry{c:16}.\\[.2em]
\bibentry{c:17}.\\[.2em]
\bibentry{c:18}.\\[.2em]
\bibentry{c:19}.\\[.2em]
\bibentry{c:20}.\\[.2em]
\bibentry{c:21}.\\[.2em]
\bibentry{c:22}.\\[.2em]
\bibentry{c:23}.\\[.2em]
\bibentry{c:24}.\\[.2em]
\bibentry{c:25}.\\[.2em]
\bibentry{c:26}.\\[.2em]
\bibentry{c:27}.\\[.2em]
\bibentry{c:28}.\\[.2em]
\bibentry{c:29}.\\[.2em]
\bibentry{c:30}.\\[.2em]
\bibentry{c:31}.\\[.2em]
\bibentry{b:1}.\\[.2em]
\bibentry{c:32}.\\[.2em]
\bibentry{c:33}.\\[.2em]
\bibentry{c:34}.\\[.2em]
\bibentry{c:35}.\\[.2em]
\bibentry{c:36}.\\[.2em]
\bibentry{c:37}.\\[.2em]
\bibentry{c:38}.\\[.2em]
\bibentry{c:39}.\\[.2em]
\bibentry{c:40}.\\[.2em]
\bibentry{c:41}.\\[.2em]
\bibentry{c:42}.\\[.2em]
\bibentry{c:43}.\\[.2em]
\bibentry{c:44}.\\[.2em]
\nobibliography{aaai22}

\end{document}